\documentclass[english]{lni}

\usepackage{booktabs}

\graphicspath{{./}}

\begin{document}

\title[PhenoEmbed]{PhenoEmbed: Self-Supervised Multispectral UAV Time-Series Embeddings for Individual Tree Crown Phenology}

\booktitle{Resilience and AI Workshop at Informatik Festival 2026}

\author[1]{Taimur Khan}{taimur.khan@ufz.de}{0000-0001-7833-5474}
\affil[1]{Helmholtz Centre for Environmental Research - UFZ\\Community Ecology\\Theodor-Lieser-Str. 4\\06120 Halle\\Germany}

\maketitle

\begin{abstract}
Tree crowns are a challenging target for resilient AI because they are not static objects: their spectral response, internal texture, translucency, and apparent boundaries change substantially across the growing season. We develop PhenoEmbed, a self-supervised crown-centric temporal embedding model trained with contrastive and masked reconstruction objectives on HeideBench, an 18-date UAV multispectral time-series benchmark for forest crown phenology in D{\"o}lauer Heide. The model treats seasonal crown dynamics as phenological appearance change driven by leaf emergence, canopy closure, senescence, and leaf-off conditions. Segmented tree crown polygons are retained as object anchors to extract aligned crown-centered crops through time, allowing one 256-dimensional vector summarizing seasonal crown appearance to be learned per tree. On 5,885 crop-safe crowns, the exported embeddings show structured low-dimensional organization, with the first two principal components explaining 25.1\% of variance and nearest-neighbor retrieval producing a median top-1 cosine similarity of 0.946. Compared with handcrafted temporal features and a learned mean-pooling baseline, PhenoEmbed yields substantially more compact nearest-neighbor structure, while ablations show that the contrastive loss, masked reconstruction loss, and explicit seasonal time features each affect the structure of the learned embedding space. These results support PhenoEmbed as a reusable forest crown representation learner and motivate future downstream tests of whether such features improve tree-level models under seasonal change.
\\
\\
\textbf{Code:} \url{https://github.com/Helmholtz-UFZ/PhenoEmbed} \\
\textbf{Model:} \url{https://doi.org/10.57967/hf/9558}\\
\textbf{Data:} \url{https://doi.pangaea.de/10.1594/PANGAEA.993969}
\end{abstract}


\section{Introduction}

\begin{figure}
\centering
\includegraphics[width=\textwidth]{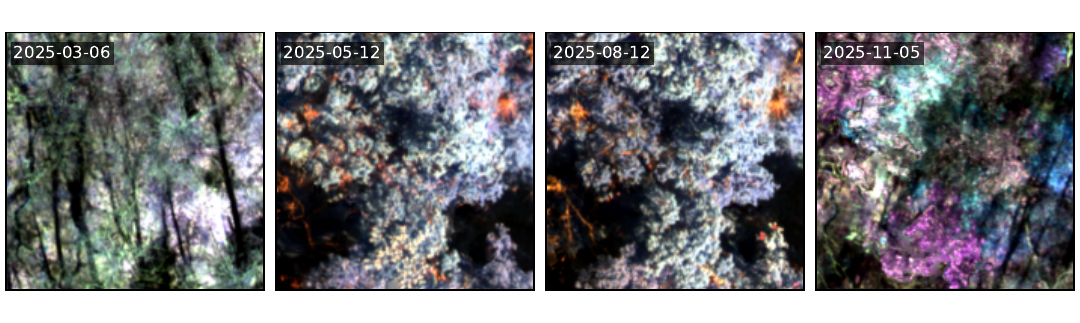}
\caption{Example crown-centered crop rendered at four points in the 2025 UAV time series. The same crown anchor is shown in early spring, green-up, late summer, and late autumn using synthetic RGB from the canonical $R$, $G$, and $RE$ bands. The panel illustrates why crowns are treated as temporally dynamic objects rather than static image patches.}
\label{fig:intro-crown-season}
\end{figure}

Tree monitoring models often assume that the object of interest has a reasonably stable visual signature. Individual tree crowns violate that assumption. Across a single growing season, the same crown can move from leaf emergence to canopy closure, from peak vigor to discoloration and senescence, and finally toward partially leaf-off conditions (\cref{fig:intro-crown-season}). Those transitions alter not only reflectance but also translucency, within-crown texture, shadowing, branch visibility, and even the apparent extent of the crown in an orthomosaic. In practice, this means that a crown segmentation or crown analysis model trained on one part of the year can become brittle when applied to another.

That brittleness is a concrete resilient AI problem in Earth observation. Season-varying remote sensing is well known to confound change analysis, because genuine structural change is entangled with phenology-induced spectral variation \cite{Kou2020SeasonalDA}. At the same time, recent individual-tree studies show that UAV imagery is rich enough to resolve phenology at the crown scale \cite{Park2019,Kleinsmann2023,Kloos2024,Kankong2026}. Embeddings are a way to compress large amounts of information into a smaller set of features that represent meaningful semantics. Therefore, the opportunity is clear: instead of treating seasonal change as nuisance variability to be suppressed, one can learn a tree representation (i.e. embeddings) that explicitly models seasonal dynamics and then reuse that representation as a prior for downstream tree-level tasks. For UAV-based forest monitoring, such representations may reduce the need for season-specific downstream models when future imagery comes from the same or similar forest environments, although transfer across sites, sensors, and years remains to be tested.

This paper presents \emph{PhenoEmbed}, a first step toward changing the usual crown-segmentation setup. The current paradigm is largely \emph{tree annotations + aerial or UAV rasters = segmentations}, supported by modern crown delineation systems \cite{Zhao2023}. This works when the target imagery resembles the imagery used for annotation and training, but it becomes fragile when the model is applied to rasters from a different phenological state. We instead aim for \emph{tree annotations + aerial or UAV rasters + phenological embeddings = more resilient segmentations}. PhenoEmbed learns one temporal descriptor per crown from a multi-date multispectral sequence, using the same broad representation-learning logic behind contrastive and masked modeling methods \cite{Chen2020SimCLR,He2022MAE,Cong2022SatMAE,Fuller2023CROMA,Wang2025FGMAE,Shenoy2024S4}. Downstream models can later use not only current image appearance but also a compact representation of how that tree changes through the season. This paper evaluates only the representation-learning part of this pipeline.

The project is framed as a reusable feature-learning system rather than a closed downstream benchmark. The implementation covers the full path from geospatial auditing and crop extraction to training, embedding export, and intrinsic representation analysis. This paper therefore has two goals:

\begin{enumerate}
\item to define the PhenoEmbed problem setting and the implemented method precisely, and
\item to report intrinsic evidence that the pipeline learns useful seasonal structure, while making clear which downstream evaluations remain future work.
\end{enumerate}

\section{Related Work}

\subsection{Individual tree crowns and UAV phenology}

Individual tree crown analysis from airborne and UAV data has advanced rapidly, with deep learning methods, reusable software, and open datasets now supporting crown delineation in high-resolution optical imagery \cite{Freudenberg2022,weinstein2020deepforest,Dubrovin2024,oamtcdpreprint,Khan2025DeepTrees}. PhenoEmbed builds on this ecosystem by treating segmented crowns as object anchors rather than only as segmentation targets. The representation unit is therefore a segmented tree crown, not a random image tile crop.

UAV phenology studies show that repeated high-resolution acquisitions can recover individual-tree seasonal signals across species, forest types, and sensors \cite{Park2019,Kleinsmann2023,Kloos2024,Kankong2026,Botticelli2026}. These results motivate the central assumption of PhenoEmbed: individual crowns carry temporal signatures that can be learned as representations, not only summarized by handcrafted vegetation indices.

\subsection{Temporal representation learning in remote sensing}

Self-supervised learning has become central to remote sensing representation learning, particularly when labeled data are limited. Masked autoencoding and contrastive learning are relevant because they naturally match multispectral and multitemporal data \cite{He2022MAE,Chen2020SimCLR}. Recent remote sensing models extend these ideas with temporal, spectral, multimodal, and feature-reconstruction targets \cite{Cong2022SatMAE,Fuller2023CROMA,Wang2025FGMAE,Shenoy2024S4}.

Temporal sequence models and geospatial foundation models further show the value of reusable representations, while also exposing the difficulty of transferring encoders across sensors, bands, scales, and tasks \cite{Cheng2023DeepSN,Xiao2024FMSurvey,Oquab2023DINOv2,Hsu2025Prithvi}. PhenoEmbed sits in a narrower regime: object-centric UAV multispectral sequences for forest crowns. It inherits lessons from foundation-model adaptation, but uses a smaller architecture suited to the available data and downstream goals.

\section{Data and Geospatial Preprocessing}

\subsection{Study data}

The experiments use HeideBench, a multispectral UAV time-series benchmark for forest crown phenology collected over a forest patch in D{\"o}lauer Heide, Halle (Saale), Germany \cite{KhanHeideBenchPangaea}. The site contains pine-dominated stands affected by dieback as well as near-natural mixed deciduous forest with oaks, birches, and beeches, making it a useful setting for observing seasonal canopy development under contrasting forest structures. The valid imaging footprint covers approximately 321,215\,m$^2$ (32.1\,ha), is bounded by 11.902653--11.911325$^\circ$E and 51.499959--51.508576$^\circ$N, and is stored in ETRS89 / UTM zone 32N (EPSG:25832). This valid-footprint polygon is used as the trusted spatial boundary for retaining crowns and materializing crops.

HeideBench provides 18 georeferenced multispectral GeoTIFF orthomosaics acquired between 6 March 2025 and 5 November 2025, spanning a 244-day seasonal period from early spring to late autumn \cite{KhanHeideBenchPangaea}. The median revisit interval is 14 days, with intervals ranging from 4 to 27 days, and the average ground sampling distance is 5.53\,cm per pixel. Data were collected with a DJI Mavic 3M Enterprise UAV equipped with multispectral cameras measuring green (560\,nm), red (650\,nm), red-edge (730\,nm), and near-infrared (860\,nm) reflectance, using RTK positioning for centimeter-level geolocation. In the GeoTIFFs, the source band order is \emph{G, R, RE, NIR, Alpha}; PhenoEmbed remaps this metadata-aware source order to the canonical modeling order \emph{R, G, RE, NIR}. 

Tree crowns are represented as 5,885 crop-safe segmented polygons included with HeideBench and extracted with the \href{https://deeptrees.de}{DeepTrees library} \cite{KhanHeideBenchPangaea,Khan2025DeepTrees}. These polygons were originally derived from RGBI aerial Digital Orthophotos (DOP) at 20\,cm resolution \cite{sachsenanhaltDigitaleOrthophotos}, merged from three overlapping DOP sources, and filtered against the valid-imaging boundary so that retained crowns lie fully within the trusted area of all 18 orthomosaics.

The preprocessing pipeline first performs a geospatial audit of the rasters and crowns. It then generates a temporal manifest that links each crown to each acquisition date together with crop geometry, date index, day-of-year, seasonal time, band mapping, and alpha-band metadata. We define a \emph{crop-safe} crown as a crown whose full 16\,m crown-centered square crop fits inside every one of the 18 rasters. Only crop-safe crowns are used for training.

Fixed-size crops are then materialized from the manifest. Each crop corresponds to a 16\,m square around the crown center, resampled to $288 \times 288$ pixels and saved as an \texttt{.npz} file containing the canonical four-band image and, optionally, the alpha mask. This representation keeps the crown polygon as the alignment anchor while producing a model-friendly tensor format. The resulting dataset contains 5,885 crowns, 18 dates, and 105,930 crop instances in total. 

\subsection{Normalization and handcrafted temporal baselines}

The model-facing dataloader normalizes crops by dividing scaled reflectance values by 10,000, clipping to a stable reflectance range, and then applying dataset-level mean and standard deviation normalization. These statistics are computed over the full crop corpus with alpha masking enabled. In parallel, the preprocessing stack computes polygon-pooled per-date summaries and common vegetation indices: Normalized Difference Vegetation Index (NDVI), Normalized Difference Red Edge (NDRE), and Green Normalized Difference Vegetation Index (GNDVI). These handcrafted trajectories serve two roles: they provide a biologically interpretable baseline, and they demonstrate the presence of seasonal signal independently of the learned embedding. \Cref{tab:data} summarizes the resulting PhenoEmbed crop-safe dataset and training corpus.

\begin{table}
\centering
\begin{tabular}{ll}
\toprule
Property & Value \\
\midrule
Orthomosaics & 18 UAV multispectral dates \\
Date range & 2025-03-06 to 2025-11-05 \\
Canonical bands & R, G, RE, NIR \\
Source band order & G, R, RE, NIR, Alpha \\
Crop-safe crowns & 5,885 \\
Temporal crop instances & 105,930 \\
Crop extent & 16\,m square per crown \\
Crop tensor size & $288 \times 288 \times 4$ \\
Train/validation split & 5,297 / 588 crowns \\
\bottomrule
\end{tabular}
\caption{PhenoEmbed dataset and training corpus.}
\label{tab:data}
\end{table}

\section{PhenoEmbed Method}

\subsection{Problem formulation}

Let crown $i$ have a temporally ordered sequence of multispectral crops
\[
X_i = \{x_{i,1}, \dots, x_{i,T}\}, \qquad x_{i,t} \in \mathbb{R}^{4 \times H \times W},
\]
where channels are in canonical order $(R, G, RE, NIR)$ and $T=18$ for this dataset. Each date also has a normalized seasonal coordinate $\tau_t \in [0,1]$ derived from day-of-year. Specifically, $\tau_t = (d_t - d_{\min})/(d_{\max} - d_{\min})$, where $d_t$ is the acquisition date and $d_{\min}$ and $d_{\max}$ are the first and last acquisition dates in the sequence. The goal is to learn an embedding function
\[
f(X_i) = e_i \in \mathbb{R}^{256},
\]
where $i$ indexes a crown, $t$ indexes an acquisition date, and $x_{i,t}$ is the four-band crop for crown $i$ at date $t$. The output $e_i$ is one vector summarizing seasonal crown appearance for downstream models.

\begin{figure}
\centering
\includegraphics[width=\textwidth]{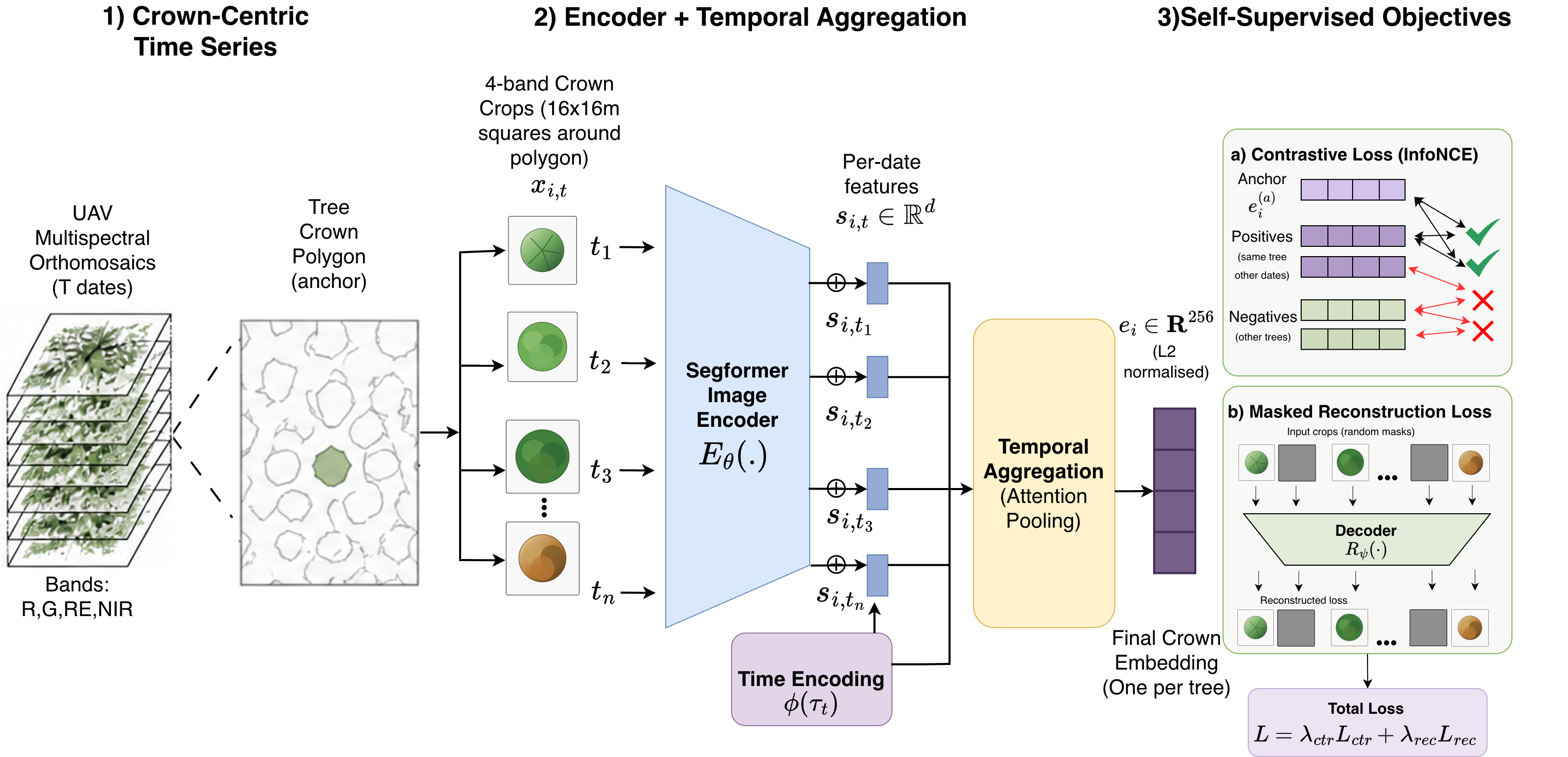}
\caption{Overview of the PhenoEmbed architecture. (1) Crown-centric time-series extraction: segmented tree crown polygons are used as object anchors to extract aligned multispectral UAV crops of the same crown across T acquisition dates, using the canonical R, G, RE, and NIR bands. (2) Encoder and temporal aggregation: each date-specific crop is passed through a shared image encoder to obtain per-date features, which are combined with seasonal time encodings and aggregated into one $L_2$-normalized 256-dimensional crown embedding. (3) Self-supervised objectives: the embedding model is trained using a contrastive loss that pulls masked views of the same crown together while separating different crowns, and a masked reconstruction loss that encourages the temporal representation to retain seasonal spectral information.}
\label{fig:phenoembed-architecture}
\end{figure}

The full PhenoEmbed architecture is summarized in \Cref{fig:phenoembed-architecture}. The model consists of three main components: (1) a crown-centric time-series extraction pipeline that produces aligned multispectral crops for each crown across all dates; (2) a spatial encoder with a multispectral adapter, followed by seasonal time encoding and temporal aggregation via a Transformer; and (3) self-supervised objectives that combine temporal contrastive learning with masked reconstruction of per-date spectral summaries. We next describe each component in more detail.

\subsection{Spatial encoder with multispectral spectral adapter}

The model uses SegFormer MiT-B2 as a shared per-date spatial encoder \cite{Xie2021SegFormer}. Because the pretrained backbone expects three-channel ImageNet-like input, we prepend a trainable $1 \times 1$ spectral adapter
\[
A: \mathbb{R}^{4 \times H \times W} \rightarrow \mathbb{R}^{3 \times H \times W}.
\]
The adapter is initialized with a physically motivated mapping: the red output channel is initialized from $R$, the green output from $G$, and the third pseudo-blue channel is initialized as the average of the $RE$ and $NIR$ bands. This initialization only defines the starting point; the $1 \times 1$ adapter is trainable, so all three pseudo-RGB channels can learn weighted mixtures of the four multispectral input bands during model fitting. This gives the pretrained backbone a reasonable starting point while allowing the mapping to adapt during training. In the selected model, the SegFormer backbone is frozen and only the adapter and temporal head are trained. This conservative choice reduces optimization instability and respects the modest size of the crown dataset.

For each date, the spatial encoder yields a pooled feature vector
\[
s_{i,t} = E(A(x_{i,t})) \in \mathbb{R}^{256}.
\]
Here, $A$ denotes the trainable multispectral adapter, $E$ denotes the shared SegFormer spatial encoder, and $s_{i,t}$ is the 256-dimensional feature vector for crown $i$ at date $t$.

\subsection{Seasonal time encoding and temporal Transformer}

For each date, the spatial encoder produces one feature vector describing the crown crop at that date. These vectors are projected into a shared feature space for the date sequence and augmented with explicit seasonal descriptors. For a normalized seasonal coordinate $\tau_t$, the implemented time feature map is
\begin{equation}
\phi(\tau_t) =
\left[
\tau_t,\;
\tau_t^2,\;
\sin(2\pi\tau_t),\;
\cos(2\pi\tau_t)
\right].
\label{eq:time}
\end{equation}
Here, $\phi(\tau_t)$ is the four-dimensional time feature vector for date $t$. These features encode both monotonic seasonal progression and cyclic annual structure. A learned linear projection maps $\phi(\tau_t)$ into the temporal token dimension and adds it to the projected spatial feature. A learned class token, used as a sequence summary, is prepended, and the resulting token sequence is processed by a 2-layer Transformer encoder with 4 attention heads and temporal dimension 256.

The output class token is finally projected and $L_2$-normalized:
\begin{equation}
e_i = \frac{W h^{\text{cls}}_i}{\|W h^{\text{cls}}_i\|_2},
\label{eq:embed}
\end{equation}
Here, $h^{\text{cls}}_i$ is the Transformer output for the class token of crown $i$, $W$ is the learned linear projection to the embedding dimension, and $e_i$ is the normalized crown embedding.

\subsection{Self-supervised objective}

Training does not require species labels or phenology labels. Instead, it uses two masked views of the same crown sequence. For each mini-batch, two independent temporal masks are sampled, and masked dates are zeroed before being passed through the encoder. This produces two embeddings $e_i^{(a)}$ and $e_i^{(b)}$ per crown.

\paragraph{Temporal contrastive loss.}
The first objective is an InfoNCE-style contrastive loss over the batch \cite{Oord2018CPC}:
\begin{equation}
\mathcal{L}_{\text{ctr}} =
-\frac{1}{B}
\sum_{i=1}^{B}
\log
\frac{\exp(\langle e_i^{(a)}, e_i^{(b)} \rangle / \gamma)}
{\sum_{j=1}^{B} \exp(\langle e_i^{(a)}, e_j^{(b)} \rangle / \gamma)},
\label{eq:contrastive}
\end{equation}
Here, $B$ is the batch size, $j$ indexes candidate crowns in the batch, $\langle\cdot,\cdot\rangle$ is cosine similarity because embeddings are normalized, and $\gamma$ is the temperature parameter. The loss is lower when two masked views of the same crown are close in embedding space and different crowns are farther apart. The selected run uses a batch size of 2 and additionally evaluate a batch-size-16 training run to examine whether a larger negative pool changes the resulting embedding structure.

\paragraph{Masked temporal reconstruction loss.}
The second objective reconstructs a compact per-date spectral target rather than raw pixels. Specifically, the target at date $t$ is the four-dimensional vector of per-band crop means:
\[
m_{i,t} = \text{mean}_{u,v}\; x_{i,t}(:,u,v).
\]
Here, $u$ and $v$ index crop pixels, and $m_{i,t}$ contains the mean $R$, $G$, $RE$, and $NIR$ values for crown $i$ at date $t$.
A linear head predicts masked-date summaries from temporal tokens, and the loss is mean squared error restricted to masked dates:
\begin{equation}
\mathcal{L}_{\text{rec}} =
\frac{1}{|\mathcal{M}|}
\sum_{(i,t)\in\mathcal{M}}
\left\|
\hat{m}_{i,t} - m_{i,t}
\right\|_2^2.
\label{eq:reconstruction}
\end{equation}
Here, $\mathcal{M}$ is the set of masked crown-date pairs and $\hat{m}_{i,t}$ is the predicted band-mean vector. This loss is lower when predicted masked-date band means match the true band means. Crown phenology is largely about how spectral state evolves through time; reconstructing band means provides a stable proxy for that temporal continuity without forcing the model to solve pixel-level synthesis.

\paragraph{Total loss.}
The total training loss is
\begin{equation}
\mathcal{L} =
\lambda_{\text{ctr}}\mathcal{L}_{\text{ctr}}
+
\lambda_{\text{rec}}\mathcal{L}_{\text{rec}},
\label{eq:total}
\end{equation}
where $\lambda_{\text{ctr}}$ and $\lambda_{\text{rec}}$ control the contrastive and reconstruction terms, with both weights set to 1.0 in the selected model. The design is closest in spirit to remote sensing methods that combine contrastive and reconstruction targets \cite{Fuller2023CROMA,Shenoy2024S4}, but it is adapted to crown-level object sequences rather than satellite tiles.

\section{Experimental Setup}

\subsection{Training configuration}

Training is implemented in PyTorch Lightning to make checkpointing, deterministic data splitting, early stopping, CSV logging, and resume behavior explicit and reproducible. Learned models are selected by minimum validation objective, and all reported embedding analyses use the selected checkpoints. The training configuration is outlined in \Cref{tab:training}.
\begin{table}
\centering
\begin{tabular}{ll}
\toprule
Component & Setting \\
\midrule
Spatial backbone & SegFormer MiT-B2 \\
Backbone initialization & ImageNet pretrained \\
Backbone status & Frozen \\
Input bands & R, G, RE, NIR \\
Spectral adapter & Trainable $1 \times 1$ conv, 4 $\rightarrow$ 3 \\
Temporal encoder & 2-layer Transformer, 4 heads \\
Embedding size & 256 \\
Batch size & 2 \\
Optimizer & AdamW \\
Learning rate & $5 \times 10^{-5}$ \\
Weight decay & $10^{-4}$ \\
Mask probability & 0.3 \\
Contrastive temperature & 0.2 \\
Validation split & 10\% of crowns \\
Checkpoint monitor & Validation objective \\
Maximum training steps & 50,000 \\
Early stopping patience & 20 validation checks \\
\bottomrule
\end{tabular}
\caption{Training configuration for the selected PhenoEmbed model.}
\label{tab:training}
\end{table}

\subsection{Evaluation protocol}

The evaluation focuses on crown-level representation quality. Learned models are selected by the best validation objective within each run, and embedding-space analyses use the selected checkpoints. The evaluation uses four complementary views:

\begin{enumerate}
\item handcrafted seasonal trajectories from polygon-pooled spectral summaries;
\item baseline and ablation comparisons against simpler representation families;
\item embedding export and analysis utilities, including PCA, nearest-neighbor retrieval, and spatial inspection;
\item a biological linear-probe analysis in which ridge regression predicts each crown's seasonal NDVI and NDRE amplitude from its full 256-dimensional embedding, evaluated using five-fold cross-validation and $R^2$.
\end{enumerate}

\subsection{Baseline and ablation protocol}

The intrinsic comparison is organized around three representation families. The first is a handcrafted temporal-feature baseline built from polygon-pooled seasonal summaries of base bands and vegetation indices. These features are standardized and evaluated with the same PCA and nearest-neighbor analysis used for learned embeddings. The second is a learned mean-pooling baseline that keeps the same SegFormer MiT-B2 spatial encoder and spectral adapter as PhenoEmbed, but replaces the temporal Transformer with simple average pooling over per-date features. The third is the full PhenoEmbed model with seasonal time encoding, temporal Transformer fusion, and the combined contrastive plus masked reconstruction objective.

The ablation plan isolates the main design choices of the temporal objective and seasonal encoding. Specifically, we evaluate variants without the contrastive loss, without the reconstruction loss, and without the explicit seasonal time features. All learned baselines and ablations use the same crop-safe crown set, train/validation split, spatial backbone, optimizer family, and checkpoint selection rule as the full model. Each completed run is exported to one embedding vector per crown and analyzed with the same PCA and cosine nearest-neighbor protocol. Validation objectives are used for checkpoint selection within runs, but not as cross-run performance metrics because the optimized objective changes across loss ablations.

We also report a batch-size-16 training-sensitivity experiment. Because this run used a different learning rate and stopping schedule, it is not treated as a controlled ablation of batch size.

\section{Results}

\subsection{Seasonal signal at crown level}

\begin{figure}
\centering
\includegraphics[width=\textwidth]{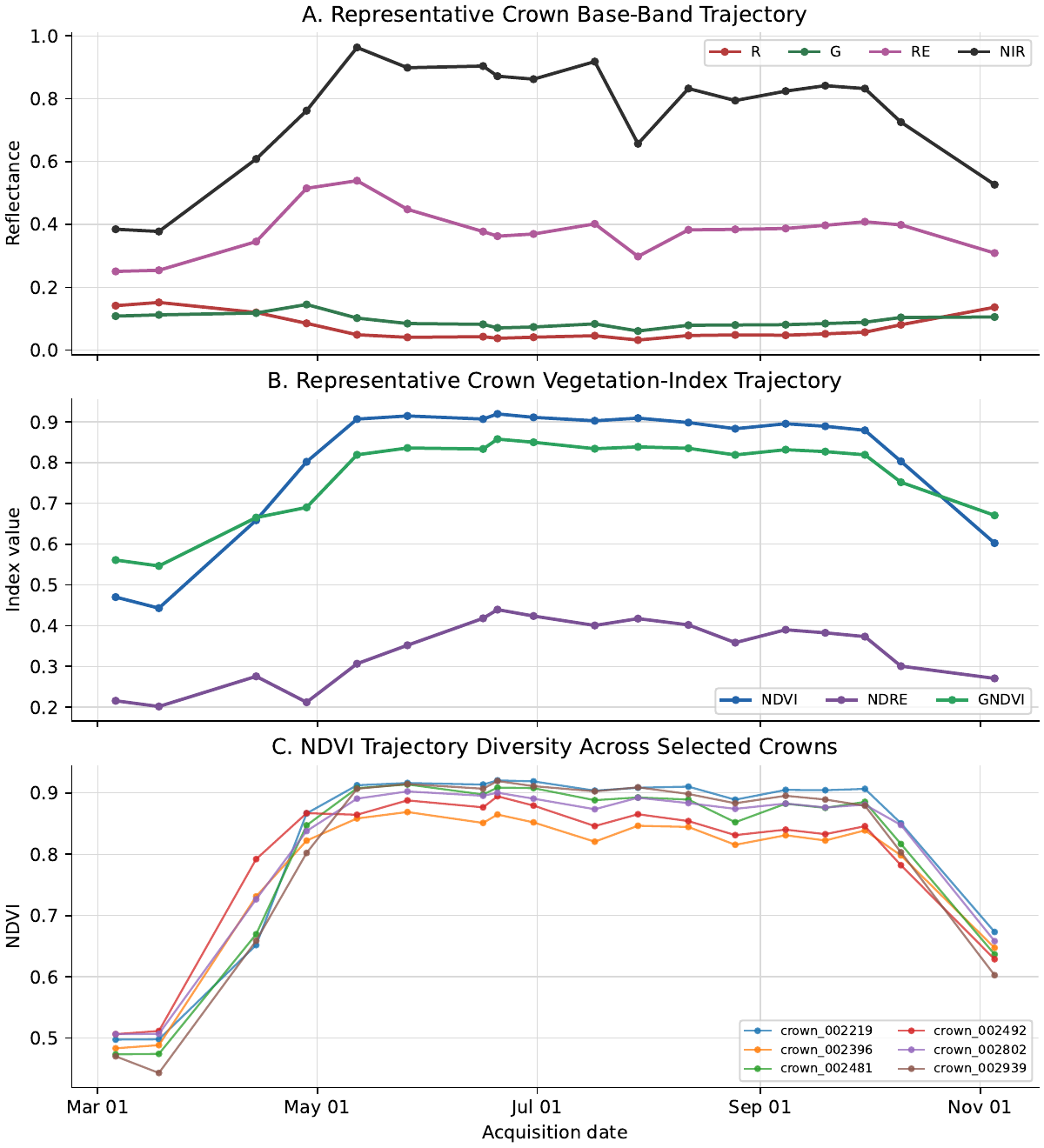}
\caption{Seasonal crown trajectories from polygon-pooled handcrafted features. The top panel shows base-band reflectance over time for a representative crown. The middle panel shows vegetation-index trajectories for the same crown. The bottom panel shows NDVI trajectories for six selected crowns with full temporal coverage.}
\label{fig:trajectories}
\end{figure}

\Cref{fig:trajectories} summarizes the handcrafted temporal trajectories extracted from polygon-pooled crown summaries. Two observations are important. First, individual crowns exhibit structured, smooth changes across the season in both base bands and vegetation indices. Second, different crowns show different amplitudes and timing, especially in NDVI-like trajectories. This is exactly the regime in which a temporal crown embedding is meaningful: the model should not only recognize a crown from one date to another, but should encode how strongly and when that crown changes.

\subsection{Selected model}

The selected full PhenoEmbed model is the checkpoint with the lowest validation objective under the configuration in \Cref{tab:training}. The contrastive term evaluates whether masked temporal views of the same crown remain close in embedding space, while the reconstruction term evaluates whether temporal tokens retain enough seasonal information to predict masked per-date spectral summaries. The exported checkpoint is summarized in \Cref{tab:selected-results}.

\begin{table}
\centering
\begin{tabular}{ll}
\toprule
Metric & Value \\
\midrule
Selection rule & Minimum validation objective \\
Exported crowns & 5,885 \\
Embedding dimension & 256 \\
PC1+PC2 variance explained & 25.1\% \\
First 8 PCs variance explained & 71.8\% \\
Median top-1 retrieval cosine similarity & 0.946 \\
\bottomrule
\end{tabular}
\caption{Selected model and embedding-space summary statistics.}
\label{tab:selected-results}
\end{table}

\subsection{Baselines and ablations}

\Cref{tab:baselines} compares the selected model with handcrafted temporal features, a learned mean-pooling temporal baseline, three ablations of the PhenoEmbed design, and a batch-size sensitivity run. The handcrafted baseline uses standardized polygon-pooled spectral and vegetation-index summaries, while the learned variants use the same crop-safe crown set and 256-dimensional embedding export pipeline as the full model.

The comparison supports three observations. First, compared with the two simpler baselines, the full temporal Transformer produces more compact nearest-neighbor structure: median top-1 cosine similarity increases from 0.855 for handcrafted temporal features and 0.847 for mean pooling to 0.946 for full PhenoEmbed. Second, removing either contrastive learning or masked reconstruction still yields compact embedding neighborhoods, with median top-1 similarities of 0.948 and 0.968. This is consistent with the contrastive branch acting mainly as a view-consistency signal in the selected batch-size-2 run. These values should not be read as a downstream ranking by themselves; they show that each individual objective can organize the embedding space, but compact retrieval neighborhoods do not prove that the representation preserves the most useful biological variation. Third, removing seasonal time encoding still retains meaningful retrieval structure but produces a shorter run and distinct PCA concentration, suggesting that explicit seasonal position affects temporal calibration.

Increasing the batch size from 2 to 16 did not improve the intrinsic embedding diagnostics under the tested training schedule: median top-1 similarity decreased from 0.946 to 0.867. Because the larger-batch run also used a different learning rate and stopping schedule, this result shows that more in-batch negatives were not sufficient to improve the representation.

We therefore interpret nearest-neighbor similarity as a diagnostic of local embedding compactness, not as an objective criterion for selecting the biologically best representation.

\begin{table}
\centering
\begin{tabular}{llrrr}
\toprule
Model & Type & Steps & PC1+PC2 & Top-1 sim. \\
\midrule
Handcrafted temporal features & baseline & -- & 37.1\% & 0.855 \\
Mean-pooling temporal encoder & baseline & 50,000 & 17.0\% & 0.847 \\
Full PhenoEmbed & baseline & 50,000 & 25.1\% & 0.946 \\
PhenoEmbed, batch 16 & sensitivity & 11,952 & 13.1\% & 0.867 \\
No contrastive loss & ablation & 50,000 & 35.3\% & 0.948 \\
No reconstruction loss & ablation & 50,000 & 30.5\% & 0.968 \\
No seasonal time encoding & ablation & 17,894 & 37.6\% & 0.929 \\
\bottomrule
\end{tabular}
\caption{Baseline, ablation, and training-sensitivity results on the crop-safe crown set. PC1+PC2 reports the percentage of embedding variance explained by the first two principal components. Top-1 sim. is reported as an intrinsic compactness diagnostic, not as a downstream performance score. Validation objectives are used only for checkpoint selection within each learned run.}
\label{tab:baselines}
\end{table}

\subsection{Embedding-space structure}

\begin{figure}
\centering
\includegraphics[width=\textwidth]{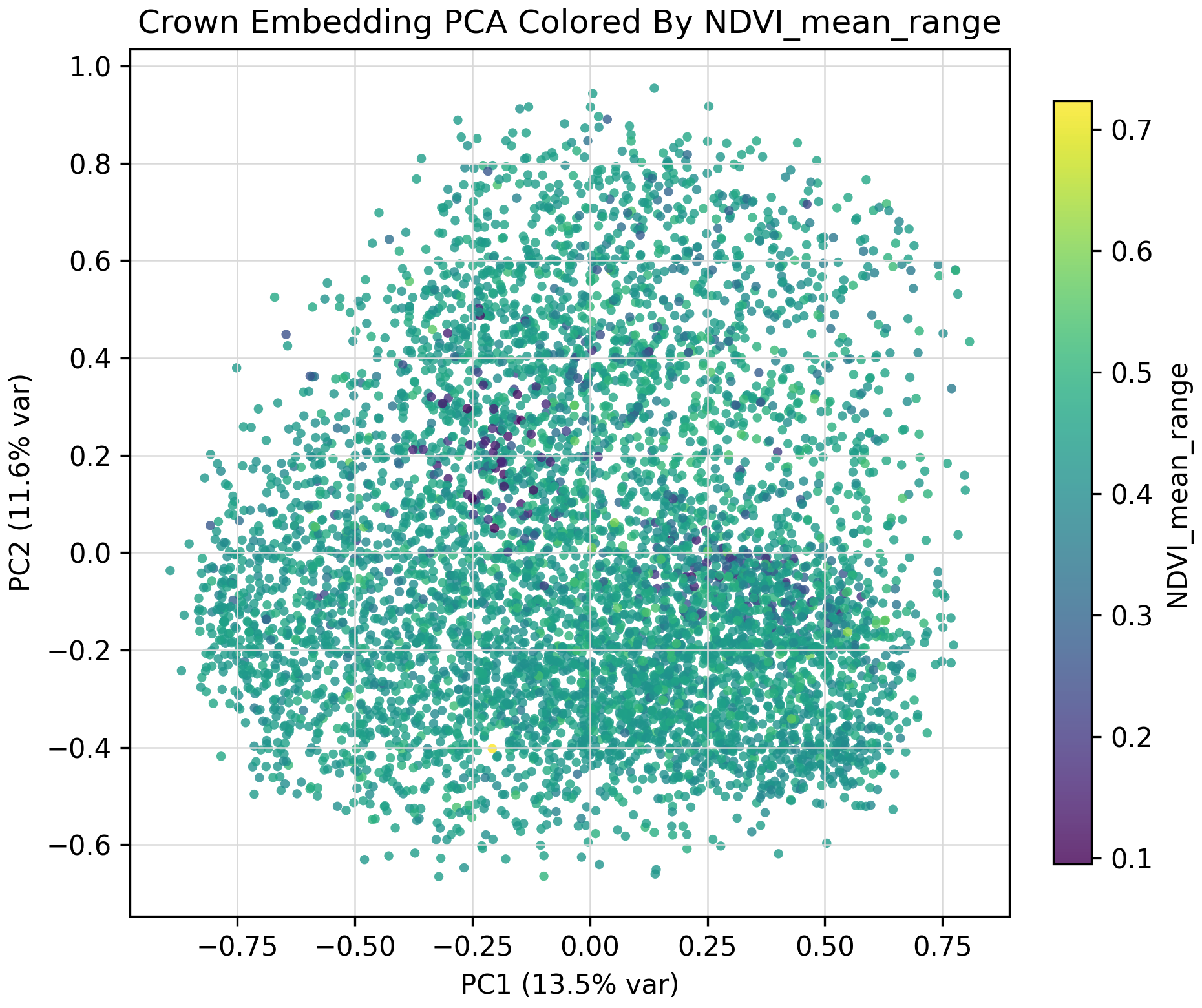}
\caption{Selected-model PCA projection colored by seasonal NDVI amplitude, defined as the maximum minus minimum polygon-pooled mean NDVI across the 18 dates. The color overlay links learned embedding axes to independently computed vegetation-index dynamics.}
\label{fig:pca-ndvi}
\end{figure}

The selected checkpoint is exported to one normalized 256-dimensional vector per crown. \Cref{fig:pca-ndvi} shows the resulting crown embedding space projected with PCA and colored by each crown's seasonal NDVI amplitude, defined as the maximum minus minimum polygon-pooled mean NDVI across the 18 dates. The first two components explain 25.1\% of the embedding variance, and the first eight components explain 71.8\%. This indicates that the model has not produced an unstructured cloud of arbitrary features. Crowns with larger seasonal NDVI amplitude tend to occupy different regions of the PCA projection than more stable crowns, suggesting that the learned embedding captures part of the seasonal vegetation signal while also retaining information beyond this single index. This comparison provides a biological sanity check: the embedding space is partly aligned with independently computed seasonal vegetation dynamics, but this does not replace downstream validation. As a second check, we tested whether the full 256-dimensional embeddings retain vegetation-index amplitude information beyond the first two PCA axes. A five-fold cross-validated ridge model predicted crown-level seasonal amplitude with $R^2=0.525$ for NDVI range and $R^2=0.414$ for NDRE range. This indicates that PhenoEmbed encodes biologically meaningful seasonal variation, while still leaving the downstream segmentation benefit to be tested directly.

\subsection{Nearest-neighbor retrieval}

Nearest-neighbor retrieval provides an additional intrinsic diagnostic of the embedding geometry. The selected model produces a median top-1 cosine similarity of 0.946, and the median cosine similarity across the top-10 retrieved neighbors is 0.902. This is not a downstream accuracy metric, but it shows that the learned representation supports crown-level search and inspection.

\begin{figure}
\centering
\includegraphics[width=0.72\textwidth]{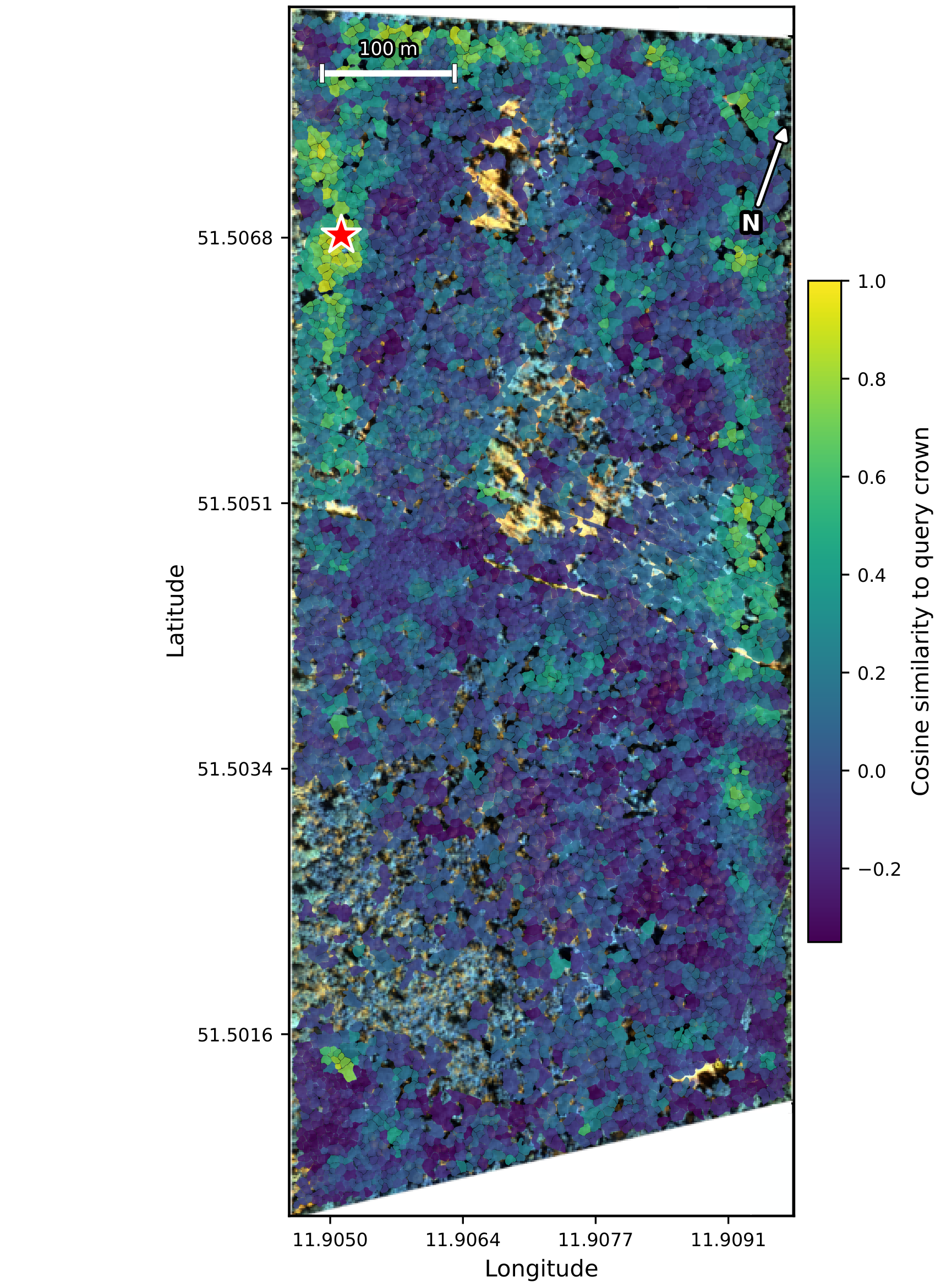}
\caption{Spatial embedding retrieval map for a representative query crown. The background is the 30 June 2025 UAV orthomosaic rendered as synthetic RGB from the canonical $R$, $G$, and $RE$ bands. The orthomosaic footprint is spatially transformed for display so that the raster fills the plotting box; the north arrow therefore follows the transformed map geometry rather than appearing vertical. Crown polygons are colored by cosine similarity between their PhenoEmbed vector and the highlighted query crown. The map provides a spatial inspection view of the learned embedding neighborhood.}
\label{fig:spatial-retrieval}
\end{figure}

\Cref{fig:spatial-retrieval} shows the same retrieval idea spatially for one representative query crown. The map is inspired by recent embedding-field views of Earth observation, where learned geospatial vectors are used as compact, queryable representations for mapping tasks \cite{Brown2025AlphaEarth}. In contrast to global satellite embedding fields, PhenoEmbed attaches one temporal vector to each annotated UAV crown object.

\section{Discussion}

\subsection{Why object-centric temporal embeddings are a good fit}

PhenoEmbed makes a strong design choice: the unit of representation is an annotated crown, not an arbitrary tile crop. This ensures the model targets trees as the primary unit of analysis. In many forestry workflows, crown annotations are already available or can be generated by a separate delineation model. Once those object anchors exist, it is much more useful to compute one robust feature vector per tree than to estimate dense motion inside a crown whose appearance is seasonally transformed. The embedding can then be attached to the crown as a persistent tree descriptor.

This perspective also aligns with the downstream goal. The project is not trying to forecast phenology for its own sake, nor to replace explicit segmentation models. It aims to produce a phenology-informed feature extractor that can later complement single-date models. A downstream model can then combine current-date appearance with a representation shaped by seasonal history.

\subsection{Limitations}

Several limitations remain. The selected model freezes the SegFormer backbone and uses a trainable four-to-three-channel adapter to reuse RGB pretraining, which is practical but compresses the multispectral signal before the backbone. The reconstruction target is also simple, using per-date band means rather than vegetation indices or spatially detailed crown structure. The selected model uses only a few in-batch negatives. A batch-size-16 sensitivity run did not improve the intrinsic diagnostics, but differences in learning rate and stopping schedule prevent a controlled attribution to batch size; a matched training comparison or queued-negative design remains future work. Most importantly, the evaluation is intrinsic: PCA, retrieval, and ablations show that the embedding space is structured, but they do not yet prove improved downstream crown segmentation under seasonal shift. Finally, the dataset is site-specific and forest-specific, so the present claim is not that PhenoEmbed is a universal geospatial foundation model, but that temporal crown embeddings are a promising route for reducing seasonal brittleness when repeated UAV multispectral coverage is available.

\subsection{Next steps}

The implementation suggests a clear roadmap:

\begin{enumerate}
\item compare the selected frozen-backbone model with a fine-tuned spatial encoder;
\item add vegetation-index auxiliary reconstruction as an ablation;
\item test whether a single-date downstream model benefits from appending PhenoEmbed features to each crown annotation;
\item use downstream transfer, rather than intrinsic retrieval alone, to choose between the full objective and the single-loss ablations.
\end{enumerate}

\section{Conclusion}

PhenoEmbed learns crown-level representations from multispectral UAV time series by using individual tree polygons as temporal anchors and training with self-supervised contrastive and masked reconstruction objectives. The results show that seasonal crown appearance contains strong structure: the learned embeddings organize crown trajectories more effectively than handcrafted spectral features and temporal mean pooling, while ablations indicate that both temporal modeling and objective design shape the representation. This supports the use of PhenoEmbed as a phenology-aware feature extractor for tree-level EO. The next step is to evaluate whether these embeddings improve downstream crown segmentation and tree analysis when training and deployment imagery come from different phenological stages.

\paragraph{LLM use declaration.}
OpenAI LLM assistance was used only to identify mistakes, check wording consistency, and make text-editing suggestions; all scientific content, experiments, interpretation, and final decisions remain the author's responsibility.

\printbibliography

\end{document}